\documentclass{article}

\usepackage{Style}

\usepackage{algorithm, caption, algpseudocode, url, hyperref, graphicx, times, latexsym, amsmath, tabularx}  %algorithm

\DeclareCaptionFormat{algor}{%
  \hrulefill\par\offinterlineskip\vskip1pt%
  \textbf{#1#2}#3\offinterlineskip\hrulefill}
\DeclareCaptionStyle{algori}{singlelinecheck=off,format=algor,labelsep=space}
\newcounter{parentalgorithm}


\usepackage{subfig}  %images
\usepackage{breqn} % berak eqs
\usepackage{multirow} % tabless 
\usepackage{hyperref}

\title{Automated Model Selection for Tabular Data}
\author{
  Avinash Amballa \thanks{Equal contribution}
  \And
  Gayathri Akkinapalli\footnotemark[1] \\
  \And
  Manas Madine 
  \And
  Priya Yarrabolu 
  \And
  Przemyslaw A. Grabowicz \\ 
  {\large College of Information and Computer Sciences}\\
  {\large  University of Massachusetts Amherst}\\
  {\large \tt \{aamballa,gakkinapalli,mmadine, nyarrabolu\}@umass.edu}
}

\date{}

\begin{document}
\maketitle

\begin{abstract}
% Structured data in the form of tabular datasets contain features that are distinct and discrete, with varying individual and relative importances to the target. Combinations of one or more features may be more predictive and meaningful than simple individual feature contributions. R's mixed effect linear models library allows for users to provide such interactive feature combinations in the model design. But in such a setting, given many features and the interactions between them to select from, model selection from the possible space of designed models becomes an exponentially difficult task. We aim to automate the model selection process for predictions on such datasets while keeping the cost of computations small. We propose a selection framework incorporating feature interactions on tabular data-sets and evaluate it on various models. Our code is available at \footnote{\url{https://github.com/AmballaAvinash/ModelSelection}}.

Structured data in the form of tabular datasets contain features that are distinct and discrete, with varying individual and relative importances to the target. Combinations of one or more features may be more predictive and meaningful than simple individual feature contributions. R's mixed effect linear models library allows users to provide such interactive feature combinations in the model design. However, given many features and possible interactions to select from, model selection becomes an exponentially difficult task. We aim to automate the model selection process for predictions on tabular datasets incorporating feature interactions while keeping computational costs small. The framework includes two distinct approaches for feature selection: a Priority-based Random Grid Search and a Greedy Search method. The Priority-based approach efficiently explores feature combinations using prior probabilities to guide the search. The Greedy method builds the solution iteratively by adding or removing features based on their impact. Experiments on synthetic demonstrate the ability to effectively capture predictive feature combinations. Code is available at \footnote{\url{https://github.com/AmballaAvinash/ModelSelection}}.

\end{abstract}

\section{Introduction}
Machine learning methods have been greatly useful at mechanising the process of dealing with input data with a large number of features. Unlike image and vision datasets, where all input components are of a similar kind and equally important, tabular datasets contain data structured into categories and continuous variables, organised into rows and columns. These datasets find applications across diverse domains, including but not limited to finance, healthcare, e-commerce. The features available in these datasets are discrete and distinct in their contributions and cannot be treated simplistically as alike unlike image/text datasets. 

Feature interactions occur when the combination of two or more features provides insights and information that cannot be derived from considering each feature in isolation. For example, the Adult Income dataset \cite{dua2019uci} contains features like \textit{education-level, marital-status, race, gender} etc., which are not equally important to income and are related in ways that might in some combinations, strongly predict income when together than treated separately. For example, it is possible that when \textit{gender=female} and \textit{marital-status=married} and \textit{age $\in$ 25-35} the combination of these factors would indicate a higher likelihood of child-bearing and maternity leave which would have a significant effect on one's income compared to combining these indicators' contribution individually: because of the way the features interact. Thus, in this context of classification and regression on such tabular data, the choice of features used plays a pivotal role in the performance of our predictors. The role of feature interactions within tabular datasets has received increasing attention in recent years in works like \cite{DBLP:journals/corr/abs-2007-14573}  and \cite{DBLP:journals/corr/abs-2008-09775}. Linear Regression (LR) models linearly aggregate features without modeling complex feature interactions. This can limit performance when inter-dependencies exist. Recognizing and effectively modeling feature interactions is pivotal in enhancing the overall performance and interpretability of models.

% \subsection{Our work}
Our work entails investigating tabular feature interactions and developing two distinct approaches for feature selection: Priority-based Random Grid Search and Greedy Search methods. We created a synthetic dataset that included categorical variables like age, education, work status, and country codes, along with interactions for marital status and gender features. A Priority-based Random Grid Search algorithm, designed for efficient exploration of feature combinations, incorporated prior probabilities to streamline the process. Greedy approach was also evaluated where we built our solution feature set iteratively. We assess feature interactions systematically while keeping in mind the computational demands.

\section{Related works}
\label{sec:rw}

\cite{DBLP:journals/corr/abs-2007-14573} introduces a framework FIVES: Feature Interaction Via Edge Search, This method involves searching for meaningful interactions among features by formulating it as a edge search over the feature graph GNN. \cite{inbook} proposed a filter-based framework to assess features, known as Relevance and Redundancy (RaR)
to incorporate multiple interactions among features and to account for redundancy in ranking features within mixed datasets. RaR generates a single score for assessing the relevance of features by taking into consideration feature interactions and redundancy.
On the other hand, \cite{10.1145/3447548.3467216} introduces the Retrieval $\&$ Interaction Machine (RIM) framework for predicting outcomes in tabular data, emphasizing both cross-row and cross-column patterns. RIM leverages search engine techniques to efficiently retrieve relevant rows and utilizes feature interaction networks for enhanced label predictions. Extensive experiments across various tasks demonstrate RIM's superiority over existing models and baselines, highlighting its effectiveness in improving prediction performance.
\\

XDeepFM \cite{DBLP:journals/corr/abs-1803-05170} adds explicit feature-product calculations to the model architecture to exploit feature interactions to improve recommender systems. They introduced the Compressed Interaction Network (CIN), which generates explicit and vector-wise feature interactions. The proposed eXtreme Deep Factorization Machine (xDeepFM) unifies CIN and DNN to learn bounded-degree and arbitrary feature interactions. \cite{ZENG20152656} introduced an interaction weight factor designed to convey whether a feature exhibits redundancy or interactivity and use this in feature selection. Though these papers do involve the idea of feature interactions, they are not applicable as baselines to our setting. This is because FIVES works with complex models on graph GNNs and XDeepFM is specific to recommender systems. In addition to this, while our task is about trying different feature combinations and automating selections from these, these works do not explore these aspects and are content with incorporating feature interactions into their respective problem setting.

\section{Background and hypotheses}

\subsection{Background}
\label{sec:the-problem}

\textbf{Feature selection:} Feature selection is a process of automatically choosing the relevant features by including relevant features or excluding irrelevant features. The aim is to enhance model efficiency and mitigate overfitting. Various strategies exist for feature selection, ranging from filter methods such as Information gain, chi-squared test that assess individual feature relevance, wrapper methods such as forward selection, backwards elimination, exhaustive feature selection that leverage model performance, to embedded methods such as regularization where feature selection is integrated into the model training process. \\ 

\textbf{Feature-relevance:} \cite{zhao2009searching} defines that a feature is relevant  only when its removal from a feature set will reduce the output confidence on a class.  Let $F$ be the full set of features, and $F_i$ be a feature, define the set $S_i = F - {F_i}$ and $P$ denote the conditional probability of class C given a feature set.  A feature $F_i$ is relevant iff $\exists S'_i \subseteq S_i$, such that $Pr(C | F_i, S'_i) \neq Pr(C |  S'_i)$. Otherwise, the feature $F_i$ is irrelevant. \\

\textbf{$k^{th}$ order feature interactions:} \cite{zhao2009searching} defines the k the order feature interactions as follows. Let $F$ be a feature subset with $k$ features $F_1, F_2, . . . , F_k$. Let $\zeta$ denote a metric
that measures the relevance of the class label with a feature or
a feature subset. We say features $F_1, F_2, . . ., F_k$ are said to interact with each other iff for an arbitrary partition $\mathcal{F} = \{\mathcal{F}_1, \mathcal{F}_2,\mathcal{F}_3, . . ., \mathcal{F}_l\}$ of $F$, where $l \geq 2$ and $\mathcal{F}_i \neq \phi$, we have  $\zeta(F) > \zeta (\mathcal{F}_i)$ $\forall$ $i \in [1,l]$ \\

\textbf{Explicit vs implicit feature capturing:} Complex models like neural networks and ensembles can implicitly capture feature interactions. This means that these models automatically learn and consider relationships between features without explicitly modeling them. While this enables powerful predictive capabilities, it often leads to a lack of explainability. Understanding how and why these models make specific predictions can be challenging due to the intricate and implicit nature of feature interactions within these models. 

In this work, we avoid the lack of explainability caused by implicit feature interactions while relying on complex models. We use linear models to explicitly model feature interactions. Linear models are inherently more interpretable, making them suitable for capturing and explaining feature relationships in a straightforward manner. We can harness the strengths of complex models for prediction while providing transparent insights into the relationships between features, ultimately yielding more explainable and trustworthy models.

\subsection{Objective and Framework} We aim to solve the feature selection problem in the specific context of tabular datasets (that have categorical features) and where many of the features we search over are interaction features formed by crossing the base features. (Note that we treat categorical values as one-hot encoded which can be crossed by set-cross product). We hypothesize that tabular datasets have features which interact in predictive manners. These interactions are not exploited with simple linear models that just add up individual base feature contributions. In addition, the space of the possible feature interactions would be too large to explore with a grid search or random search in terms of computational costs. 

\begin{figure}[t]
    \centering
    \includegraphics[width=0.8\linewidth]{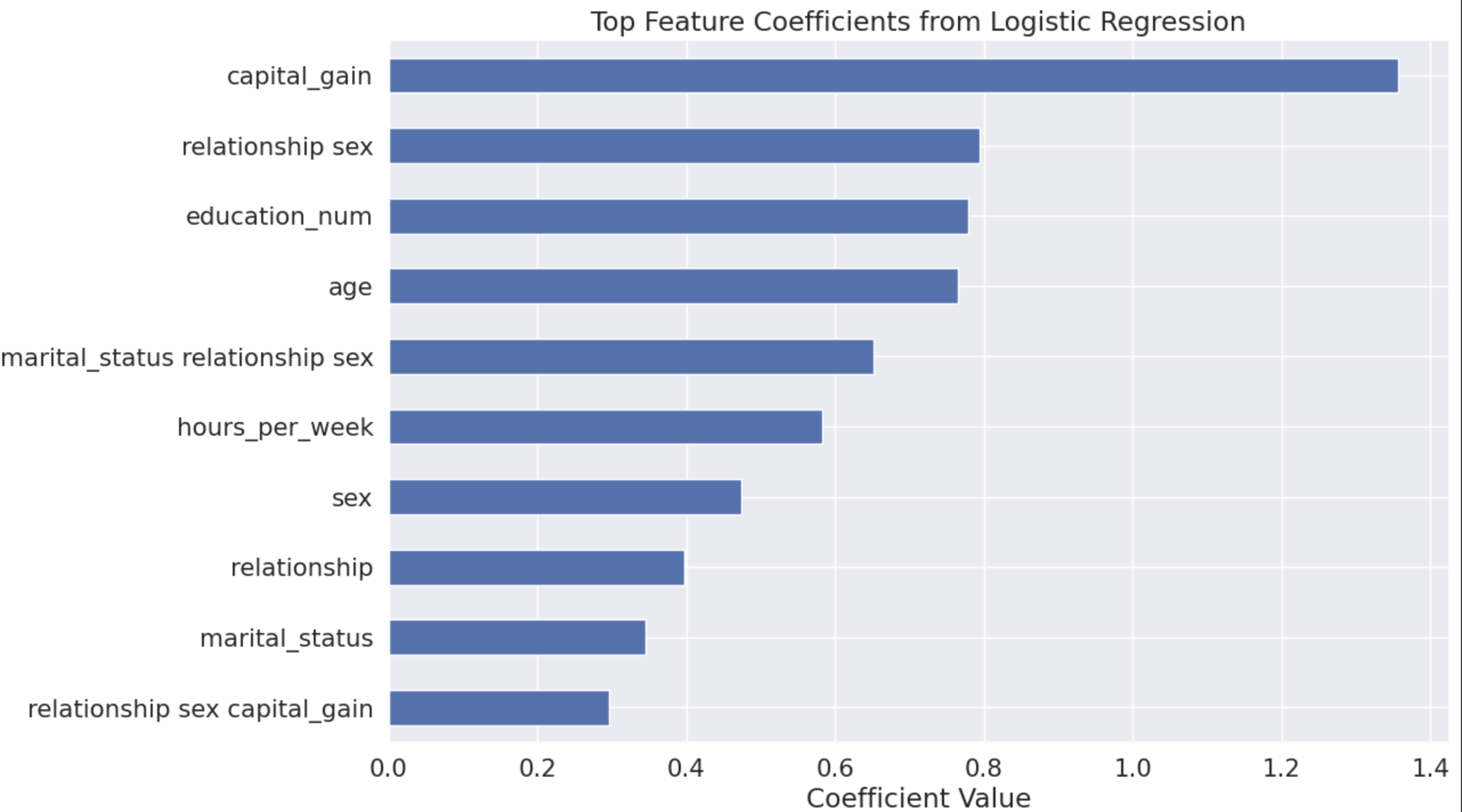}
    \caption{Feature importance's in the real-life Adult Income dataset.}
    \label{fig:feat-importance-adult}
\end{figure}

We see the potential for automating this search for an optimum model, while keeping computational costs small. We optimise our search using prior assumptions on the extent of feature interactions and using feature importance's to cut our search space short.  We optimize the metrics of the model such as $R^2$ value (coefficient of determination), MSE (mean squared error), Log-likelihood and AIC (Akaike Information Criterion) to show the performance of our models.  In addition, we track runtime of feature search to evaluate computational efficiency of the search. We use the following frameworks for out experiments: \textit{sklearn}, \textit{numpy}, \textit{statsmodels OLS} and \textit{pandas}. Since, we work with linear regression models applied on categorical variables, we apply \textit{sklearn}'s LinearRegression functionality upon various one-hot encoded pandas dataframes. We encode categorical features using one-hot encoding before fitting linear regression models.

\section{Data}

\subsection{US Adult Income dataset and discussion}

We conducted experiments on the US Adult Income dataset \cite{dua2019uci}, aiming to predict income levels (above or below $50k$) using details about the person such as their age, education, occupation status, race, gender etc. Our focus was on assessing the impact of feature interactions on model performance. The pipeline integrated sklearn's PolynomialFeatures module to generate interaction terms and a Logistic Regression classifier, coupled with hyperparameter tuning via grid search.

\textbf{Grid search on Adult Income Data}: We use Sklearn's PolynomialFeatures module with the \texttt{interactions\_only} parameter set to true, eliminating squared terms. GridSearchCV optimized the logistic regression model, considering interactions via PolynomialFeatures. The search spanned various degrees of polynomial features, ensuring the selection of the optimal model variant in terms of accuracy, balancing complexity with predictive power. Post GridSearchCV, `capital\_gain' emerged as the most significant predictor. Notable interaction terms included `relationship-sex' and `maritalstatus-relationship-sex', highlighting their relevance. The feature importance plot is depicted in Figure \ref{fig:feat-importance-adult}.

% Degrees 2 and 3 were also considered, generating expanded feature sets with original and interaction terms to model interactions.
% For instance, with original features \(A\), \(B\), and \(C\), a 2-degree polynomial expansion would create features \(A, B, C, AB, AC, \) and \(BC\), while a 3-degree expansion would further include \(A, B, C, AC, AB , BC\) and \(ABC\). 

% The grid search revealed optimal performance with a regularization parameter of 0.01, l2 regularization, and a polynomial function degree of 3. 
This interaction-based model achieved a test accuracy of 0.7911, while the original linear model (trained without interactions) yielded a test accuracy of 0.817 for Logistic Regression. If interactions in the dataset were significant, we would expect higher accuracy with our interaction-based model. We thus hypothesize that our assumption of significant interactions in this real-life dataset did not hold true.

\begin{figure}[t]
    \centering
    \includegraphics[width=0.5\linewidth]{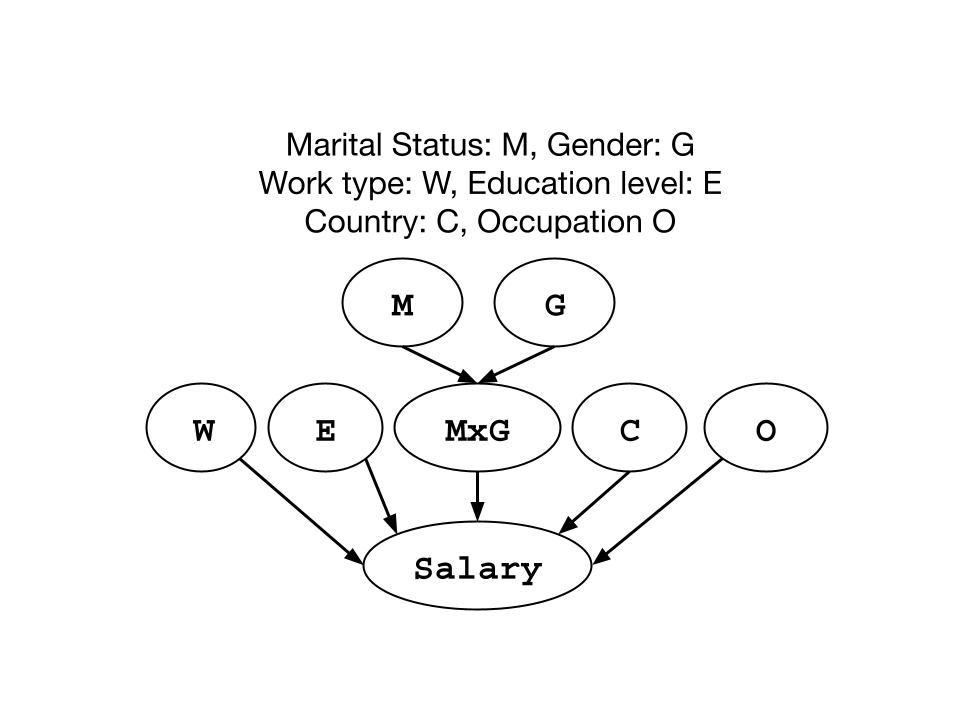}
    \caption{Causal diagram showing the interactions between the features used to create the synthetic dataset.}
    \label{fig:causaldigaram}
\end{figure}

\begin{figure*}[t]
            \centering
            \includegraphics[width=0.9\textwidth]{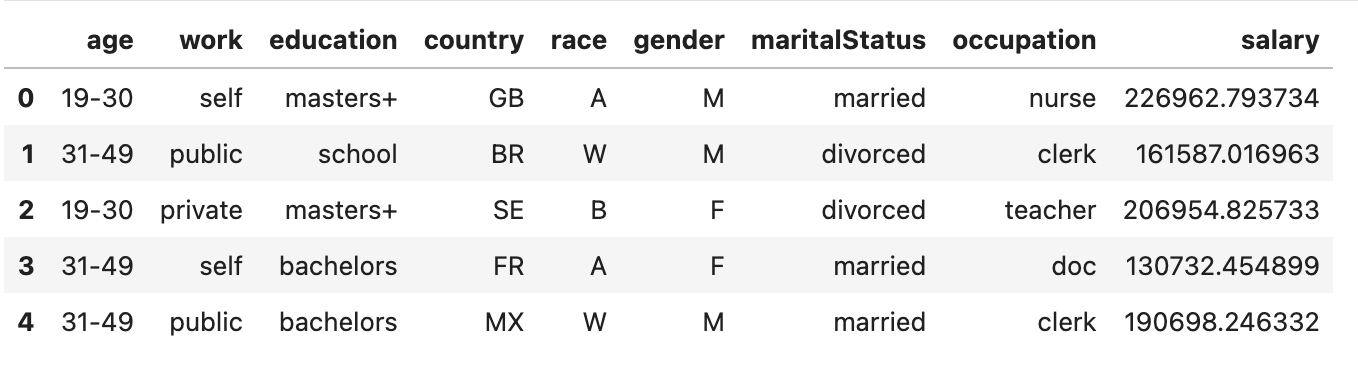}
            \caption{A snippet of the generated synthetic dataset Adult2.}
            \label{fig:dataset}
        \end{figure*}

\begin{figure*}[t]
            \centering
            \includegraphics[width=0.9\textwidth]{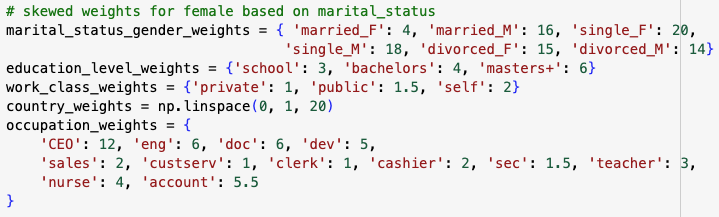}
            \caption{Weight assignment in Adult2 creation}
            \label{fig:datagen}
        \end{figure*}

\subsection{Synthetic data generation}
\label{sec:syn-data}
 Based on the above analysis on the Adult dataset, we aim to create a synthetic dataset where we incorporate some desired feature interactions. These feature interactions serve as the ground truth to evaluate our method on. We generated a synthetic dataset we call `Adult2', deliberately incorporating feature interactions. We assigned weights to the features and combinations of features as illustrated in figure \ref{fig:datagen}. These weights were then used to combine the contributions of the features into the target (label) attribute, salary.

Formally, let $F$ be the full set of features and $F_I \subseteq F$ be the subset of features with 2nd order features (interactions) i.e., $F_I = \{(F_i,F_j)\}$ where $ F_i, F_j$ are the features with interactions. Define $F' = F - F_I$ (to avoid features being redundant by appearing both in first and second order forms). We define true label $y = f(F', F_I)$. In the above example, we take the function $f$ to be linear.

 The core of our dataset creation process is our self-assigned weights (figure \ref{fig:datagen}) for each of the  ground truth model's features. We convert the given data's feature set into a final feature set by crossing-among and selecting features. We then add up their individual contributions to create our target column.  Note: incorporation of feature interactions into our dataset generation above is done by assigning weights to crossed-feature combinations - which we call our interaction features. These interaction features would be non-linear in terms of the individual components. Our dataset has categorical features such as `Age', `Work', `Education', `Country', `Race', `Gender', `MaritalStatus', and `Occupation'. In the data generation phase, we assign weights to these features and combinations of these features, so as to combine their contributions into the target (label) attribute - salary.

We generate data attributes by uniform random generation over the possible value space. In our setting, we introduced interactions via a designed feature, `MaritalStatusGender' to capture the interaction between `MaritalStatus' and `Gender'. We set the weights for this feature so that it is not linear in terms of the individual components (see fig. \ref{fig:datagen}). This feature, along with a subset of base features is weighted and added up to generate our target attribute `Salary'. This process is illustrated in figure \ref{fig:causaldigaram}. Let all the features used be called ($F_{used}$). At the end, to introduce variability, we inject a noise factor to Salary, creating a dataset with 2000 samples. A snippet from this dataset is displayed in figure \ref{fig:dataset}. \\

\textbf{Data Analysis:} We conducted training on two linear regression models using different datasets. The first model, trained on the raw data (X, y) using just the first order features ($F$) provided directly without interactions, yielded an $R^2$ value of 0.70. We then sanity-checked our interaction generation by manually creating a second model, which uses the same features and interactions as the true hidden model ($F_{used}$) we used to create the data. This model achieved a 0.9957 $R^2$ score showing the effectiveness of an interaction based feature set. \footnote{$R^2$ value serves as an indicator of the extent to which the model explains the variability in the target variable y (salary). Higher $R^2$ values imply a better-fitting model.}

This shows that these features without interactions are not sufficient to explain the dataset variance under a linear model: we needed to explore the crossed-feature interactions to get the full performance. Thus in our dataset, the target/true label y is better explained as a function of input feature interactions than features without interactions.

\section{Techniques and methods}

\subsection{Computational complexity of the feature space}
Initially, we considered a full grid search, exploring nearly $2^{n + {n \choose 2} + {n \choose 3} + ... + {n \choose n}}$ potential combinations of features and interactions, where n represents the number of features in the dataset, and ${n \choose r}$ represents the interactions of degree r. Due to the computational complexity and substantial computation time associated with all these possibilities, we \textit{decided to restrict to exploring interactions only via the second order}. This still leaves $2^{n + {n \choose 2}}$ possible feature subsets involving the base features and the 1st order interaction features. 

\subsection{Priority-based random grid search}

So we had to simplify it further by assuming that \textit{there would be very few feature interactions} among the ${n \choose 2}$ combinations involved in the true model. We took the number of 2nd order feature interaction terms as being [0, 1, 2].

\begin{algorithm}[t]
\caption{Priority based random grid search}\label{alg:priority}
\begin{algorithmic}[1]
\Require $D$ (dataset with $n$ features), $N$ (number of iterations)
\State Let $F = \{f_1, f_2, ..., f_n\}$ (base features)
\State Let $I = \{f_{jk} | 1 \leq j < k \leq n\}$ (2$^{nd}$ order interaction features)
\State Init: $F_{\text{best}} \gets \emptyset$, $R_{\text{best}} \gets 0$.

\For{$iteration = 1$ to $N$}
\State Sample $k \sim \text{Prior}(k)$, for \#interactions
\State Sample $I_{\text{selected}} \subset I$, where $|I_{\text{selected}}| = k$
\State Sample $F_{\text{sampled}} \subset F$, ensuring already selected interaction features are excluded.
\State $F_{\text{cur}} = F_{\text{sampled}} \cup I_{\text{selected}}$ (to remove redundant features)
\State Build LR linear model($F_{\text{cur}}$), train for $R^2$ score $R_{\text{cur}}$
\If{$R_{\text{cur}} > R_{\text{best}}$}
\State $F_{\text{best}} = F_{\text{current}}$, $R_{\text{best}} = R_{\text{cur}}$

\EndIf
\EndFor

\State \Return $F_{\text{best}}$
\end{algorithmic}
\end{algorithm}

Thus our feature selection process (refer algorithm \ref{alg:priority}) could be done via separate selections over the base and interaction feature set spaces: 1. A small number k from the ${n \choose 2}$ combinations' set and 2. Most of the features from the base feature columns. For 1., we first sample k, using \textit{assigned prior probabilities on k ranging from 0 to 2}. Then these k interaction features are randomly sampled following the prior distribution from the combination set. To ensure that there exist no redundancy in the features among both the selections and we remove the features that are in the sampled interaction feature set (first selection set) from the overall base features before sampling the set of features without interaction. In 2., base feature set is randomly sampled from the remaining features. The three sections italicized above are the priors we have imposed on our search to make it simpler and easier, in contrast to a naive brute-force grid search over features. We expect these priors to give us decent results with lesser number of iterations.

In summary, our priority search approach divides the naive random grid search into separate prior-based grid searches across base features and interaction features to better reach the optimum feature set.

\begin{algorithm}[t]
\caption{Backward Elimination}\label{Algo-BE}
\begin{algorithmic}[1]
\Require Dataset $D$ (X,y), Significance level $\alpha$
    \State Encode categorical features of $X$ including 2nd-order interactions into n One-Hot features
    \State Init: $F_{\text{selected}} \gets  \{f_1, f_2, ..., f_n\} $
    \State Fit a linear model with ($F_{\text{selected}}, y$), store p-values and std error for all features
    \State $pvalue_{\text{max}} \gets$ maximum p-value
    \While{$pvalue_{\text{max}} > \alpha$}
        \State $f_{\text{worst}} \gets$ feature with highest p-value and standard error
         \State Remove $f_{\text{worst}}$ from $F_{\text{selected}}$
        \State Fit a linear model with ($F_{\text{selected}}, y$) and store p-values and std error for all features
    \State $pvalue_{\text{max}} \gets$ maximum p-value
    \If{abrupt increase in AIC}
            \State \textbf{break}
        \EndIf
    \EndWhile
    \State \Return $F_{\text{selected}}$
\end{algorithmic}
\end{algorithm}

\begin{algorithm}[t]
\caption{Forward Selection}\label{Algo-FS}
\begin{algorithmic}[1]
    \Require Dataset $D$ (X,y), Significance level $\alpha$
    \State Encode categorical features of $X$ including 2nd-order interactions into n One-Hot features
    \State Let $F = \{f_1, f_2, ..., f_n\}$ (individual features)
    \State Init: $F_{\text{selected}} \gets \emptyset$
    \State $pvalue_{\text{min}} \gets$ -$\infty$
    \While{$pvalue_{\text{min}} < \alpha$}
        \State Fit a linear model with ($F_{\text{selected}} \cup f_i, y$). Repeat this $\forall$ $f_i \in F$, and store \{$f_i$: (pvalue for feature $F_i$ , std error for feature $F_i$ )\} across all the models
        \State $f_{\text{best}} \gets$ feature with minimal p-value and std error 
        \State $pvalue_{\text{min}} \gets$ minimal p-value
        \State Add $f_{\text{best}}$ to $F_{\text{selected}}$
        \State Remove $f_{\text{best}}$ from $F$
        \If{abrupt increase in AIC}
            \State \textbf{break}
        \EndIf
    \EndWhile
    \State \Return $F_{\text{selected}}$
\end{algorithmic}
\end{algorithm}

\subsection{Greedy search}

%Given the large feature space, a bottom-up feature building approach is well-suited for our data.

We explore the approaches to feature selection using a greedy strategy. Greedy search is a heuristic optimization strategy that makes locally optimal choices at each step with the hope of finding a global optimum. In the context of feature selection in machine learning, greedy search involves iteratively making decisions to add or remove features based on their immediate impact, aiming to improve the overall model performance. Backward Elimination and Forward Selection are two variants of greedy search used for feature selection. \\  

\textbf{Backward Elimination:} Described in algorithm \ref{Algo-BE}, this top down approach involves fitting the model with all possible features initially, considering both interactions and individual features. It then iteratively removes the least important features, narrowing down the feature set for optimal model performance. To identify features that have a lesser impact on the model, our strategy involves assessing p-values, and standard errors with respect to each feature. The feature with higher p-value and standard error with respect to the fitted linear model are removed. Since our data contains only categorical features we apply one hot encoding before fitting the model.

\begin{figure*}[t]
    \centering
    \includegraphics[width=\linewidth]{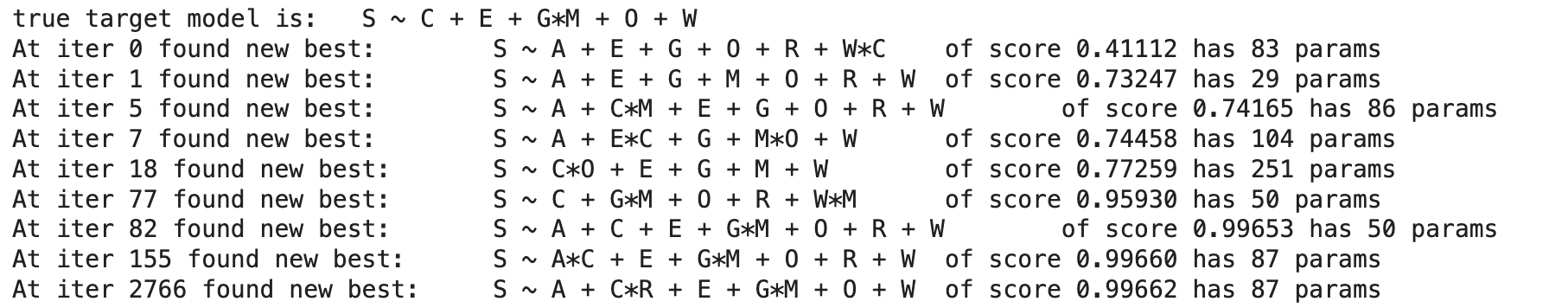}
    \caption{Evolution of the best feature set over iterations of the priority search algorithm. From the figure, we observe that the algorithm reached near optimal model}
    \label{fig:evolution}
\end{figure*}

\textbf{Forward Selection:} Described in algorithm \ref{Algo-FS}, this bottom-up approach entails creating individual simple regression models for each feature in the dataset. p-values and standard errors are then calculated for these models, and the feature with the lowest p-value and standard error is selected. This chosen feature is added to models of other features, resulting in one less linear regression model, but each of them will have two features. Continue iterating the approach keeping track of the selected features until the lowest p-value from a model is no longer below the significance level, typically $\alpha$ =  0.01 \footnote{p-value $<$ 0.01, we reject the null hypothesis}. This approach refines the model by gradually including features with significant contributions.

\section{Results}

\subsection{Priority-based random grid search}

Figure \ref{fig:evolution} shows our result for priority search. We observe that the algorithm achieved the near optimal model ($\sim A+C*R+E+G*M+O+W$) after 2766 iterations. We also achieve an optimal $R^2$ of 0.996 because the synthetic data was made using the same linear additions we capture in our LR models. 

\begin{figure}
    \centering
    \includegraphics[width=0.8\linewidth]{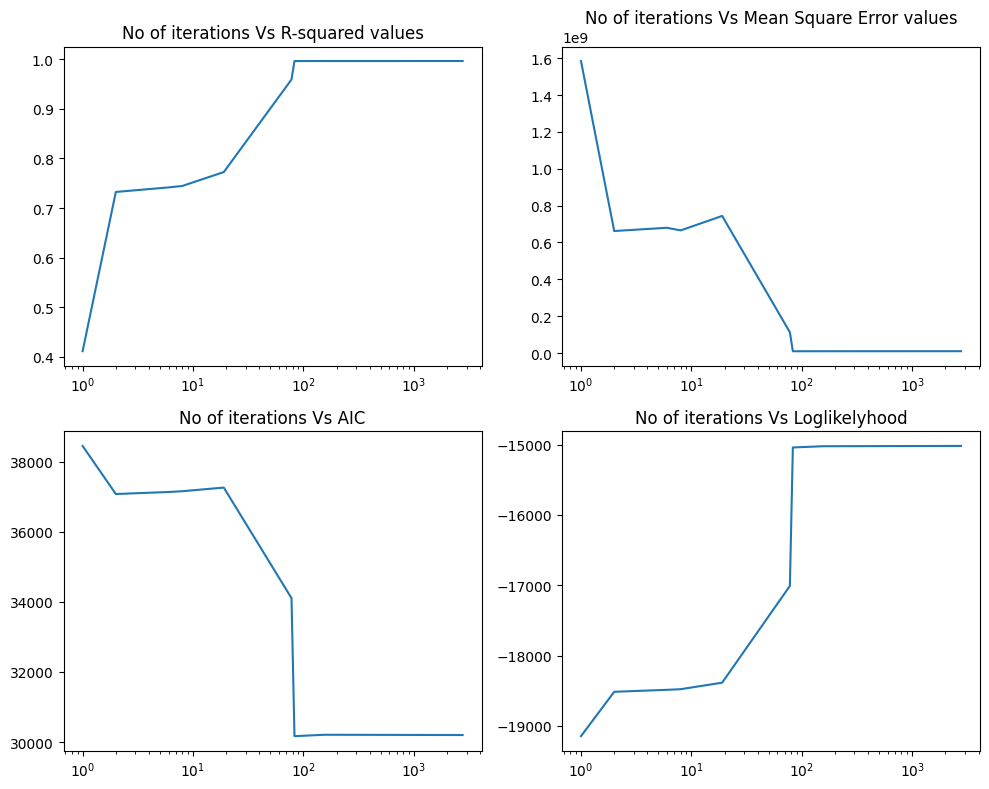}
    \caption{Priority search performance. The x-axis employs a logarithmic scale.}
    \label{fig:iters-r2}
\end{figure}

Figure \ref{fig:iters-r2} depicts the $R^2$, MSE, log likelyhood and AIC value of the best model identified in relation to the number of times the algorithm was executed. On average, the peak performance of the best model is observed around the $100^{th}$ iteration. Total iterations for all possible feature subsets for our dataset (under highest degree of interaction being 2) would've been $2^{8 + {8 \choose 2}}$. 

% In the next steps we plan to properly investigate the evolution of the best feature set over iterations and track how the spikes in $R^2$ occur closely.

\subsection{Greedy Search}
\begin{figure}[t]
    \centering
    \includegraphics[width=0.8\linewidth]{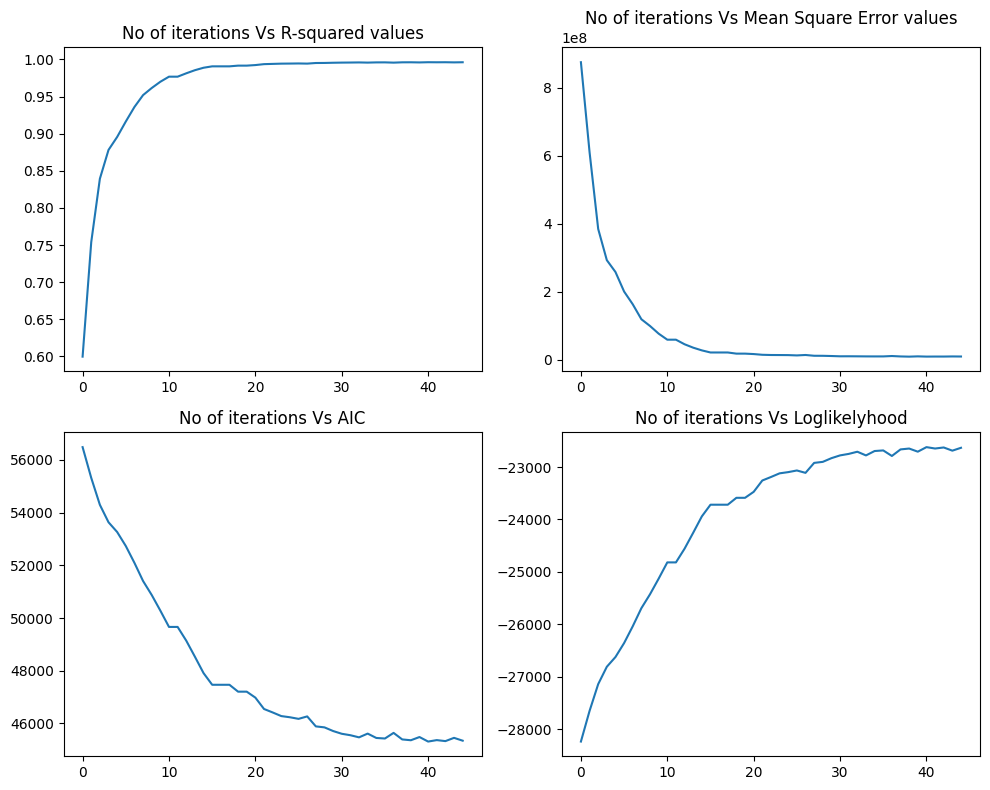}
    \caption{Forward Selection performance. We terminated forward selection when the p-value of the added feature increased beyond 0.01, which implies that an insignificant feature has been added.}
    \label{fig:forward_selection}
\end{figure}

\begin{figure}[t]
    \centering
    \includegraphics[width=0.8\linewidth]{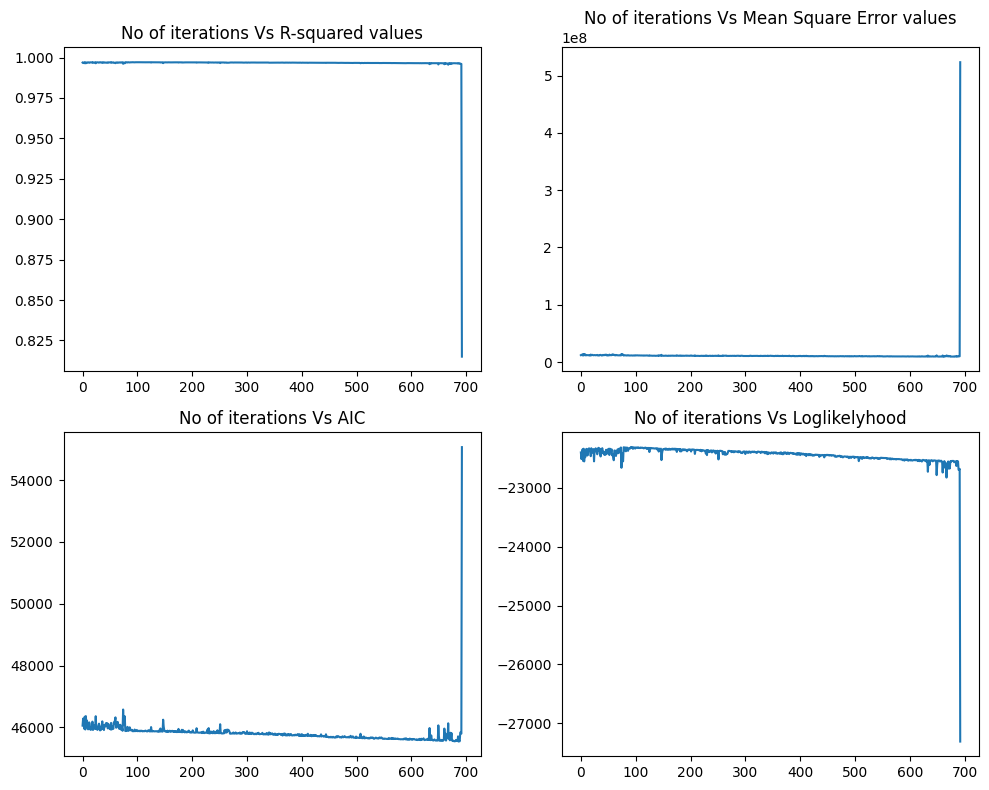}
    \caption{Backward Elimination performance. We terminate backward elimination due to sudden surge in AIC, which implies that a crucial features has been removed. Debugging this revealed that the removed feature is $M*G$ which is a crucial interaction feature in the true model. Hence, we backtrack one step to restore this feature and terminate the algorithm.}
    \label{fig:backwardelimination}
\end{figure}

 Considering the 8 base features in the synthetic data (Figure \ref{fig:dataset}), and the 2nd order interactions features (28), there are 36 features and the potential feature space after one-hot encoding is 953 (coefficients). This results in a total of $2^{953}$ feature subsets, making it computationally intensive to find the best model. Thus we use greedy approach to optimize the search space. Figure {\ref{fig:forward_selection}}, \ref{fig:backwardelimination} shows the learning curve for Forward selection algorithm and backward elimination respectively considering the metric. 
 
\subsection{Discussion}
 Table \ref{tab:Table 1} summarizes the results from priority search, backward elimination and forward selection. From the table \ref{tab:Table 1}, we can conclude that the optimal solution using priority-based grid search, forward selection and backward elimination may differ. \\

For the synthetic dataset we performed all the above mentioned methods and observed the following results:

\begin{enumerate}
    \item Random Priority search achieved a 0.9966 $R^2$ with an average runtime of 2 to 3 minutes, capturing a near-optimal (near true) model. This method required fewer iterations and took less time.

    \item The Forward selection achieved a 0.9962 $R^2$  on the synthetic data with an average runtime of 20 to 30 minutes, capturing the true model.

    \item  The Backward Selection achieved over 0.9968 $R^2$  with an average runtime of 10 to 15 minutes, capturing a near-optimal solution.

\end{enumerate}

From the experiments, the forward selection seems to give us the true model when compared to the other approaches.  Priority based random grid search required the fewer computational runtime when compared with the other greedy approaches. This highlights a trade-off between computational efficiency and performance. Importantly, all methods outperformed baseline linear models (models without interactions), demonstrating the value of modeling feature interactions. 

 \begin{table*}[t!]  % to place at top of page
     \centering
      \begin{footnotesize}  
     \begin{tabular}{ cccc }
     \hline
      \textbf{Algorithm} &  \textbf{Priority based-random grid search} & \textbf{Forward Selection} & \textbf{Backward Elimination} \\
     \hline
       Obtained Model & S $\sim$  A + C + E + M*G + O + W*R  & S $\sim$  C + E + M*G + O + W &  S $\sim$  A*R + C*O +  E + M*G + W\\
      Avg. runtime on COLAB & 1-3 min & 20-30 min & 10-15 min \\
      $R^2$ & 0.9966 &  0.9962 & 0.9968 \\
      % MSE & 9589705.77 & 10480472.75 & 9879365.33\\
      % AIC & - & 45682.66 & 45582.69\\
      % Log-Likelihood & - & -22801.33 & -22534.34 \\
      MSE & 1.01e7 & 8.83e6 & 1.09e7\\
      AIC & 3.01e4 & 4.53e4 & 4.57e4\\
      Log-Likelihood & -1.5e4 & -2.26e4 & -2.24e4 \\

     \hline
     \end{tabular}
     \caption{Comparing priority search, backward elimination and forward selection statistics. We also report the generated model against the true model, which is S $\sim$ C + E + M*G + O + W. We use google colab with T4 GPU runtime with 12GB RAM to run the experiments.}
     \label{tab:Table 1}
     \end{footnotesize}
 \end{table*}

\section{Conclusion}
% \begin{enumerate}

    % \item Improve the \textbf{Priority based Random Grid} search by learning the prior probability distribution on k, also by increasing the size of the k. We plan to downweigh the probability by the feature size, that is to down sample the features that are large for example: country x occupation gives 20 x 12 = 240 possible values.
    % \item Improve the \textbf{Greedy algorithm} considering the H-statistics.
    % \item Explore \textbf{regularization} methods for feature selection. 
    % \item SHAP analysis on forward selection vs backward elimination
In this work, we investigated automated model selection incorporating feature interactions for tabular datasets. We hypothesized that combinations of features can be more predictive than individual contributions. We proposed three approaches: a Priority-based Random Grid Search, a Greedy Search method with forward selection, and backward elimination. The Priority search uses priors to efficiently explore feature subsets. The Greedy approach iteratively adds/removes features based on impact. Experiments conducted on real world Adult dataset reveals that there are no significant interactions hence we fall back to create synthetic dataset on which forward selection algorithms achieves true model whereas priority-based random grid search achieves near optimal model with minimal computational time. 

\section{Future work} 
% The first area of improvement involves refining the Priority based Random Grid search by incorporating a learned prior probability distribution on k. Additionally, expanding the size of k and implementing downweighting based on feature size is planned, with the aim of addressing large feature spaces such as country x occupation, which may result in a substantial number of possible values. Another avenue for improvement is the optimization of the Greedy algorithm by considering H-statistics. The exploration of various regularization methods for feature selection is also slated for investigation. Lastly, a comparative SHAP analysis between forward selection and backward elimination will be conducted to gain insights into the interpretability and impact of these feature selection approaches. These proposed enhancements collectively aim to advance the effectiveness and efficiency of the current research methodology.

In future work, there are several areas for enhancement in our research methodology. Firstly, we plan to refine the Priority-based Random Grid search by integrating a learned prior probability distribution on parameter k. Additionally, we intend to extend the range of parameter k and introduce down-weighting based on feature size to better accommodate large feature spaces such as \textit{country $\times$ occupation}, which may contain numerous possible values. Finally, we aim to conduct a comparative SHAP analysis between forward selection and backward elimination to assess the interpretability and impact of these feature selection approaches. These proposed enhancements collectively seek to enhance the effectiveness and efficiency of our current research methodology.

% Another avenue for improvement involves optimizing the Greedy algorithm by incorporating H-statistics. Furthermore, we will explore various regularization methods for feature selection. 
% \end{enumerate}

\newpage
\bibliographystyle{apalike}
\footnotesize
\bibliography{yourbib}

\end{document}